\newif\ifpaper
\newif\ifarxiv
\newif\ifedits

\paperfalse \arxivtrue \editstrue 

\documentclass[letterpaper]{article} 
\usepackage{aaai24}  
\usepackage{times}  
\usepackage{helvet}  
\usepackage{courier}  
\usepackage[hyphens]{url}  
\usepackage{graphicx} 
\usepackage{natbib}  
\usepackage{caption} 
\usepackage{algorithm}
\usepackage{algpseudocode}

\usepackage{newfloat}
\usepackage{listings}
\DeclareCaptionStyle{ruled}{labelfont=normalfont,labelsep=colon,strut=off} 
\floatstyle{ruled}
\newfloat{listing}{tb}{lst}{}
\floatname{listing}{Listing}
\usepackage{amsfonts}
\usepackage{amsthm}
\usepackage{amsmath}
\usepackage{amssymb}
\usepackage{amsfonts}
\usepackage{mathtools}
\usepackage{bm}

\usepackage{tikz}
\usepackage{forest}
\usetikzlibrary{automata, positioning}
\usetikzlibrary{matrix,intersections,arrows,decorations.pathmorphing,backgrounds,positioning,fit,petri}
\usetikzlibrary{decorations.pathmorphing} 
\usetikzlibrary{fit}					
\usetikzlibrary{backgrounds}	
\usepackage{float}

\tikzset{%
  zeroarrow/.style = {-stealth,dashed},
  onearrow/.style = {-stealth,solid},
  c/.style = {circle,draw,solid,minimum width=2em,
        minimum height=2em},
  r/.style = {rectangle,draw,solid,minimum width=2em,
        minimum height=2em}
}

\newcommand {\bigO}{\mathcal{O}}
\newcommand {\St}{\mathcal{S}}
\newcommand {\Obs}{\mathcal{O}}

\newcommand {\A}{\mathcal{A}}

\newcommand {\Hist}{\mathcal{H}}
\newcommand {\Dist}{\mathcal{D}}
\newcommand {\Reward}{\mathcal{R}}

\newcommand {\R}{\mathbb{R}}
\newcommand {\E}{\mathbb{E}}
\newcommand {\N}{\mathbb{N}}

\newcommand{\comp}{\, \raisebox{0.1em}{$\scriptscriptstyle \circ$} \,}

\newcommand{\abs}[1]{\lvert #1 \rvert}
\newcommand{\barmu}{\bar{\mu}}
\newcommand{\vbar}{\;\middle|\;}

\newcommand{\sS}{\mathsf{S}}
\newcommand{\sE}{\mathsf{E}}
\newcommand{\sI}{\mathsf{I}}
\newcommand{\sR}{\mathsf{R}}

\newcommand{\DoNothing}{\mathit{DoNothing}}
\newcommand{\Vaccinate}{\mathit{Vaccinate}}
\newcommand{\Quarantine}{\mathit{Quarantine}}

\newtheorem{theorem}{Theorem}
\theoremstyle{definition}
\newtheorem{defn}{Definition}

\newtheorem{prop}{Proposition}

\newtheorem*{T1}{Theorem \ref{thm:DHA_value_mu}}
\newtheorem*{T2}{Theorem \ref{thm:DHA_value_constant}}

\title{Dynamic Knowledge Injection for AIXI Agents}
\author{
    Samuel Yang-Zhao\textsuperscript{\rm 1},
    Kee Siong Ng\textsuperscript{\rm 1},
    Marcus Hutter\textsuperscript{\rm 1, 2}
}
\affiliations{
    \textsuperscript{\rm 1}Australian National University\\
    \textsuperscript{\rm 2}Google DeepMind\\
    samuel.yang-zhao@anu.edu.au, keesiong.ng@anu.edu.au,  http://www.hutter1.net/
    
}

\begin{document}

\maketitle

\begin{abstract}
Prior approximations of AIXI, a Bayesian optimality notion for general reinforcement learning, can only approximate AIXI's Bayesian environment model using an a-priori defined set of models. This is a fundamental source of epistemic uncertainty for the agent in settings where the existence of systematic bias in the predefined model class cannot be resolved by simply collecting more data from the environment. We address this issue in the context of Human-AI teaming by considering a setup where additional knowledge for the agent in the form of new candidate models arrives from a human operator in an online fashion. We introduce a new agent called DynamicHedgeAIXI that maintains an exact Bayesian mixture over dynamically changing sets of models via a time-adaptive prior constructed from a variant of the Hedge algorithm. The DynamicHedgeAIXI agent is the richest direct approximation of AIXI known to date and comes with good performance guarantees. Experimental results on epidemic control on contact networks validates the agent's practical utility.
\end{abstract}

\section{Introduction}
The AIXI agent \cite{Hutter:04uaibook} is a Bayesian solution to the general reinforcement learning problem that combines Solomonoff induction \cite{solomonoff97} with sequential decision theory. At time $t$, after observing the action-observation-reward sequence $h_{1:t-1} \coloneqq aor_{1:t-1}$, the AIXI agent computes the action $a_t$ to choose via an expectimax search up to a horizon $H$ with a mixture model:

\begin{multline}
    a_t = \arg\max_{a_t} \sum_{or_t} \max_{a_{t+1}} \sum_{or_{t+1}} \ldots \max_{a_{t+H}} \sum_{or_{t+H}} \\
    \left[ \sum_{j = t}^{t+H} r_j \right] \sum_{\rho \in M_U} 2^{-K(\rho)} \rho(or_{1:t+H} | a_{1:t+H}). \label{eqn:aixi1}
\end{multline}
The Bayesian mixture ${\sum\limits_{\rho \in M_U} 2^{-K(\rho)} \rho(or_{1:t+H} | a_{1:t+H})}$ is AIXI's environment model and is computed as a mixture over the set $M_U$ of all enumerable chronological semi-measures with each element $\rho \in M_U$ assigned a prior according to its Kolmogorov complexity $K(\rho)$. Note that $M_U$ is formally equivalent to the set of all computable distributions; from this perspective, AIXI can be considered the ultimate Bayesian reinforcement learning agent. Furthermore, \cite{Hutter:04uaibook} shows that AIXI's environment model converges rapidly to the true environment and its policy is pareto optimal and self-optimising. 

The AIXI agent can be viewed as containing all possible knowledge as its Bayesian mixture is performed over all computable distributions. From this perspective, AIXI's performance does not suffer due to limitations in its modelling capacity.
In contrast, all previous approximations of AIXI are limited to having a finite pre-defined model class containing a subset of computable probability distributions, presenting an irreducible source of error.
To address this issue, we introduce \textit{dynamic knowledge injection}, a setting where an external source is used to provide additional knowledge that is then integrated into new candidate environment models. 
In particular, dynamic knowledge injection models human-AI teaming constructs where the human can provide additional domain knowledge that the agent can use to model aspects of the environment. 
Once a new environment model is proposed, the central issue is then to determine how it can be incorporated to improve the agent's performance. Utilising a variation of the GrowingHedge algorithm \cite{MM17:growing_expert}, itself an extension of Hedge \cite{cesa2006prediction}, we construct an adaptive \textit{anytime} Bayesian mixture algorithm that incorporates newly arriving models and also allows the removal of existing models.
DynamicHedgeAIXI is the richest direct approximation of AIXI to date and comes with strong value-convergence guarantees against the best available environment sequence.
We validate the agent's performance empirically on multiple experimental domains, including the control of epidemics on large contact networks, and our results demonstrate that DynamicHedgeAIXI is able to quickly adapt to new knowledge that improves its performance. 

\subsection{Related Work.}  
While AIXI is only asymptotically computable, it serves as a strong guiding principle in the design of general purpose AI agents. \cite{veness09} gives the first tractable approximation of AIXI by using the Context Tree Weighting algorithm (CTW) \cite{WST95} to restrict the Bayesian mixture to a set of variable-order Markov environments and Monte-Carlo Tree Search to approximate the expectimax operation.
This was followed by a body of work on extending the Bayesian mixture learning to larger classes of history-based models \cite{venessNHB12},\cite{venessWBG13}, \cite{bellemareVB13}, \cite{bellemareVT14}.
The current best approximation of AIXI is given in \cite{yang-zhao2022a}, which introduces the $\Phi$-AIXI-CTW agent to extend AIXI's approximation to non-Markovian and structured environments. This is achieved using the $\Phi$-BCTW data structure, a model that extends the CTW algorithm's representation capacity by combining state abstraction, motivated by the Feature Reinforcement Learning framework \cite{Hutter2009FeatureRL}, with a rich logical formalism. One limitation of $\Phi$-AIXI-CTW is that it models the environment using state abstractions that are fixed after performing feature selection at the start. The current work alleviates this issue by extending $\Phi$-AIXI-CTW to the dynamic knowledge injection setting.

\section{Background and Notation}

\subsection{General Reinforcement Learning}
We consider finite action, observation and reward spaces denoted by $\A, \Obs, \Reward$ respectively. The agent interacts with the environment in cycles: at any time, the agent chooses an action from $\A$ and the environment returns an observation and reward from $\Obs$ and $\Reward$. We will denote a string $x_1x_2\ldots x_n$ of length $n$ by $x_{1:n}$ and its length $n-1$ prefix as $x_{<n}$. An action, observation and reward from the same time step will be denoted $aor_t$. A history $h$ is an element of the history space $\Hist \coloneqq (\A \times \Obs \times \Reward)^* $. An environment $\rho$ is a sequence of probability distributions $\{\rho_0, \rho_1, \rho_2, \ldots \}$, where $\rho_n: \A^{n} \to \Dist((\Obs \times \Reward)^n)$, that satisfies $\forall a_{1:n}~ \forall or_{<n} \; \rho_{n-1}(or_{<n} | a_{<n}) = \sum_{or \in \Obs \times \Reward} \rho_n(or_{1:n} | a_{1:n})$. We will drop the subscript on $\rho_n$ when the context is clear. The predictive probability of the next percept given history and a current action is given by $\rho(or_n | aor_{<n}, a_n) = \rho(or_n | h_{n-1}, a_n) \coloneqq \frac{\rho(or_{1:n} | a_{1:n})}{\rho(or_{<n} | a_{<n})}$ for all $aor_{1:n}$ such that $\rho(or_{<n} | a_{<n}) > 0$. 

The general reinforcement learning problem (GRL) is for the agent to learn a \textit{policy} $\pi: \Hist \to \Dist(A)$ mapping histories to a distribution on possible actions that will allow it to maximise its future expected reward. In this paper, we consider the future expected reward up to a finite horizon $H \in \N$. 
Given the history $h_{<t} = aor_{<t}$ up to time $t$ and a policy $\pi$ the value function with respect to the environment $\rho$ is given by $V_{\rho}^{\pi}(h_{<t}) \coloneqq \E_{\rho}^{\pi} \left[  \sum_{i = t}^{t+H} r_i \vbar h_{<t}  \right]$, where $r_i$ denotes the random variable distributed according to $\rho$ for the reward at time $i$. The action value function is defined similarly as $Q_{\rho}^{\pi}(h_{<t}, a_{t}) \coloneqq \E_{\rho}^{\pi} \left[  \sum_{i = t}^{t+H} r_i \vbar h_{<t}, a_{t}  \right]$. The agent's goal is to learn the optimal policy $\pi^*$, which is the policy that results in the value function with the maximum reward for any given history.

\subsection{Abstract Environment Models}
In the dynamic knowledge injection setting, the candidate models received by the agent may only approximate the underlying environment. 
State abstractions provides a framework for defining such models.
A state abstraction is a mapping $\phi: \Hist \to \St_{\phi}$ that maps the space of history sequences into an abstract state space.
For the history $h_{1:t}$ at time $t$, the state at time $t$ is given by $s_t = \phi(h_{1:t})$.
In this manner, the interaction sequence of the original process is mapped to a state-action-reward sequence.
For a given $\phi$, an abstract Markov Decision Process (MDP) predicts the next state and reward according to a distribution $\rho_\phi: \St_{\phi} \times \A \to \Dist(\St_{\phi} \times \Reward)$ that factorises into a state transition and reward distribution as $\rho_\phi(s', r | s, a) = \rho_{\phi}(s' | s, a) \rho_{\phi}(r | s, a, s')$. 
Let $\rho_{\phi} \coloneqq (\rho_{t, \phi})_{t \geq 1}$, where $\rho_{t, \phi}: \St_{\phi} \times \A \to \Dist(\St_{\phi} \times \Reward)$ for all $t$. 
We refer to $(\phi, \rho_{\phi})$ together as an abstract environment model. 
Note that the state of an abstract MDP model will in general not be a sufficient statistic for the underlying environment at hand and thus presents a source of bias; a simple example is when $\phi$ maps all histories into a single state.

An abstract MDP can help simplify the environment's dynamics but pushes a lot of the complexity into the design of the state abstraction function and a sufficiently powerful representation is required to ensure as little generality is lost. 
In particular, the quality of an abstract environment model will determine how closely its reward distribution approximates the underlying environment's reward distribution.
Following \cite{yang-zhao2022a}, we consider the class of predicate environment models, which are abstract environment models that can be constructed from a set of predicate functions on histories.
More precisely, we consider abstract environment models where the state abstraction is of the form $\phi(h) = (p_1(h), \ldots, p_n(h))$ and $p_i: \Hist \to \{0, 1\}$ are predicates. Under a sufficiently powerful knowledge representation and reasoning language (such as the one described in \cite{lloyd11,lloyd03,farmer}), such models are capable of representing a large class of non-Markovian and structured environments as abstract MDPs. 
The $\Phi$-BCTW data structure, described in the next section, will be used to model the distributions under the predicate state abstractions.

\subsection{Bayesian Mixtures and $\Phi$-BCTW}
The importance of Bayesian mixture models in general reinforcement learning is that they converge rapidly to the `good' models in the model class. 

\begin{theorem}\cite{Hutter:04uaibook}
	\label{thm:MEMConvergence}
	Let $\mu$ be the true environment and $\xi$ be the mixture environment model over a model class $\mathcal{M}$.
	Let $x_k = or_k$ and $h_{<k} = aor_{<k}$. For all $n \in \N$ and for all $a_{1:n}$,
	\begin{multline}
	\sum_{k=1}^{n} \sum_{x_{1:k}} \mu(x_{<k} | a_{<k}) \left( \mu(x_k | h_{<k} a_k) - \xi(x_k | h_{<k}a_k) \right)^2 \\ 
        \leq \min_{\rho \in \mathcal{M}} \left\{ \ln \frac{1}{w^\rho_0} + KL(\mu || \rho) \right\} \label{eqn:MEMBound}
	\end{multline}
\end{theorem}

Context Tree Weighting (CTW) \cite{WST95} is a rare case where a Bayesian mixture over prediction suffix trees can be computed efficiently. 
The $\Phi$-BCTW data structure introduced in \cite{yang-zhao2022a} generalises CTW by allowing predicates $p_i : \Hist \to \{0,1\}$ from a set $\Phi$ to act as the internal nodes of the tree. 
In more details, in the $\Phi$-BCTW tree, each sub-tree of depth $d$ is a $\Phi$-prediction suffix tree ($\Phi$-PST) model. For a history $h$, a $\Phi$-PST with predicates $p_i$ at depth $i$ computes a path from root to leaf node as $p_1(h)p_2(h)\ldots p_d(h)$, which forms a state abstraction. At each leaf node resides a KT estimator \cite{KT06} maintaining a distribution over the next bit. 
By chaining together multiple $\Phi$-BCTW trees, the data structure can be used to predict the binary representation of arbitrary symbols.

A $\Phi$-BCTW data structure constructed using a set $\Phi$ of $D$ predicates is able to perform an exact Bayesian mixture over $2^{(2^D)}$ $\Phi$-PST models in $\bigO(D)$ time.
Since each $\Phi$-PST model represents a different predicate environment model, applying $\Phi$-BCTW to predicting state and reward symbols then amounts to computing a mixture environment model over all predicate environment models that can be constructed from $\Phi$. The following result states the environment mixture result for $\Phi$-BCTW.
\begin{prop}[\cite{yang-zhao2022a}]
    \label{prop:pbctw_mixture}
    Suppose a state and reward symbol can be binarized in $k$ bits. Let $\mathcal{T}_d$ be the set of $\Phi$-PST models that can be constructed from a single $\Phi$-BCTW tree of depth $d$. 
    Let $P(\cdot | a_{1:n}, T)$ denote the action-conditional distribution under model $T$.
    Then the $\Phi$-BCTW computes a mixture environment model of the form, where $\Gamma(T)$ is the coding length of $T$:
    \begin{align}
        \label{eqn:phibctw_mixture}
        \xi(sr_{1:n} | a_{1:n}) = \sum_{T \in \mathcal{T}_{d} \times \ldots \times \mathcal{T}_{d+k-1}} 2^{-\Gamma(T)} 
        P(sr_{1:n} | a_{1:n}, T)~.
    \end{align}
\end{prop}

\subsection{Prediction with Expert Advice}
The prediction with expert advice setting is a well-established framework providing theoretically sound strategies on how to aggregate the forecasts provided by many experts in a sequential setting \cite{cesa2006prediction}. This setting is characterised by a game played between a learner and an adversary. Initially, a loss function $\ell: \mathcal{X} \times \mathcal{Y} \to \R$ is provided where $\mathcal{X}$ is the vector space of predictions and $\mathcal{Y}$ is the outcome space. The learner has access to a set of fixed experts $M$. At time $t$, a learner receives prediction $x_{t, i} \in \mathcal{X}$ from expert $i$. The learner then must combine the predictions from all experts and outputs $x_t \in \mathcal{X}$. An adversary then chooses an outcome $y_t \in \mathcal{Y}$ causing the learner to incur loss $\ell_t = \ell(x_t, y_t)$ and observe the loss $\ell_{t, i} = \ell(x_{t, i}, y_t)$ for each expert $i$. Learners are typically designed to minimise the \textit{regret} $L_T - L_{T, i} = \sum_{t=1}^{T} \ell_{t} - \sum_{t=1}^{T} \ell_{t, i}$, a measure of the relative performance of the agent with respect to any fixed expert $i \in M$. 
The \textit{Hedge} (exponential weights) algorithm is a simple yet fundamental algorithm 
in this setting \cite{cesa2006prediction, Vovk98}. Given a prior distribution $\bm{\nu}$ over $M$ and $\eta > 0$, Hedge predicts 
\begin{align*}
    x_t = \frac{\sum_{i \in M} w_{t, i} x_{t, i}}{\sum_{i \in M} w_{t, i}}
\end{align*}
where $w_{t, i} = \nu_i e^{- \eta L_{t-1, i}}$. The weights of the Hedge algorithm can be viewed as the posterior probabilities of each expert \cite{Jordan1995WhyTL}. The following is a standard regret bound for the Hedge algorithm.

\begin{prop}[\cite{cesa2006prediction}]
\label{prop:hedge_regret}
If the loss function $\ell$ is $\eta$-exp-concave, then for any $i \in M$, Hedge with prior $\bm{\nu}$ has regret bound $L_T - L_{T, i} \leq \frac{1}{\eta} \log \frac{1}{\nu_i}.$
\end{prop}

\section{Dynamic Knowledge Injection}
In this section we formalise the Dynamic Knowledge Injection setting. We first present an extension of the prediction with expert advice setting known as the \textit{specialists} setting. The Dynamic Knowledge Injection setting can then be naturally described using the specialists framework.

\subsection{The Specialists Setting}
Incorporating expert advice from novel experts arriving in an online fashion can be cast into the specialists setting \cite{FSSW97}. The specialist setting extends the prediction with expert advice setting by introducing specialists: experts that can abstain from prediction at any given time step. In this setting, the learner has access to a set $M$ of specialists where at time $t$, only specialists in a subset $M_t \subseteq M$ output predictions $x_{t, i} \in \mathcal{X}$. 
The crucial idea to adapt the Hedge algorithm to this setting was presented in \cite{CV09} where inactive specialists $j \notin M_t$ are attributed a forecast equal to that of the learner.
More precisely, choosing 
$x_{t, j} = \frac{ \sum_{i \in M_t} w_{t, i} x_{t, i} }{ \sum_{i \in M_t} w_{t, i} }$
for  $j \notin M_t$ results in $$x_t = \frac{\sum_{i \in M} w_{t, i} x_{t, i}}{\sum_{i \in M} w_{t,i}} = \frac{ \sum_{i \in M_t} w_{t, i} x_{t, i} }{ \sum_{i \in M_t} w_{t, i} }.$$

The \textit{specialist aggregation} algorithm uses the above `abstention' trick where the weight for expert $i$ at time $t$ is given by $w_{t, i} = \nu_i e^{-\eta L_{t-1, i}}$ and $L_{t, i} \coloneqq \sum_{s \leq t: i \in M_s} \ell_{s, i} + \sum_{s \leq t: i \notin M_s} \ell_s$. 
This abstention trick helps maintain enough weight on abstaining experts to ensure the regret is well controlled. We use this fundamental technique to incorporate newly arriving models for Bayesian agents.

\subsection{The Dynamic Knowledge Injection Setting}
The Dynamic Knowledge Injection setting can now be naturally defined using the Specialists framework.
In this setting, a specialist $i$ is an abstract environment model of the form $(\phi_i, (\rho_{t, i})_{t \geq 1})$ where $\phi_i$ is a state abstraction function constructed from a set of predicate functions.
As an abstract MDP, $(\rho_{t, i})_{t \geq 1}$ is a sequence of distributions where $\rho_{t, i}: \St_{i} \times \A \to \Dist(\St_{i} \times \Reward)$.
Modelling the environment as a sequence allows us to naturally represent environment models that update and learn their distributions online, such as $\Phi$-BCTW. 
Over the course of the agent's lifetime, it is given new abstract environment models by a human operator at intermittent time steps.
A key example of this setting is when the initial models our agent starts with are inadequate and a separate training process is able to submit new and improved models over time.

\subsection{Incorporating New Models}
The key issue for the agent is to determine how the newly arriving models can be integrated in an online fashion.
The GrowingHedge algorithm \cite{MM17:growing_expert} applies to the setting where the set of active experts grows over time and achieves the same regret bound as specialist aggregation. 
In practice, the growing experts setting is infeasible as computational constraints dictate that the set of available experts cannot grow unboundedly over time. Instead, we consider the setting where arriving experts are \textit{contiguous specialists} that
can only be active for contiguous periods of time before becoming inactive forever. 
This captures the situation in practice whereby the set of models can grow until resource limits are reached and a newly entering model must then replace an older model.
Formally, over $T$ steps the set $T_i = \{t \in [T]: i \in M_t\}$ for any contiguous specialist $i$ is a contiguous set of integers. Let $\tau_i = \min(T_i)$ and $\kappa_i = \max(T_i)$, representing the arrival and `death' times for specialist $i$.

\begin{figure}[t]
\begin{algorithm}[H]
\caption{DynamicHedge (modifies GrowingHedge \cite{MM17:growing_expert})}\label{alg:dynamic_hedge}
\begin{algorithmic}[1]
\State \textbf{Require:} Learning rate $\eta > 0$, weights $\bm{\nu} = (\nu_i)_{i \geq 1}$, sequence of sets of contiguous specialists $(M_t)_{t \geq 1}$.
\State \textbf{Initialize:} $L_0 = 0$. For $i \in M_1$, set $w_{1, i} = \nu_i$.
\For{$t = 1, 2, \ldots, T$}
\State For all ${i \in M_t}$, receive prediction $x_{t, i} \in \mathcal{X}$.
\State Predict
   $ x_t = \frac{ \sum_{i \in M_t} w_{t, i} x_{t, i} }{ \sum_{i \in M_t} w_{t, i} }$.
\State Observe $y_t \in \mathcal{Y}$.
\State Set $\ell_t = \ell(x_t, y_t)$ and $\ell_{t, i} = \ell(x_{t, i}, y_t)$.
\State Set $L_t = L_{t-1} + \ell_t$.
\State For $i \in M_t \cap M_{t+1}$, set $w_{t+1, i} = w_{t, i} e^{-\eta \ell_{t, i}}$. \label{alg:DH_update1}
\State For $i \in M_{t+1} \setminus M_t$ set $w_{t+1, i} = \nu_i e^{-\eta L_t}$. \label{alg:DH_update2} 
\EndFor
\end{algorithmic}
\end{algorithm}
\end{figure}
Under this restriction we present the DynamicHedge algorithm (Algorithm \ref{alg:dynamic_hedge}).
DynamicHedge modifies GrowingHedge's weight update to account for contiguous specialists; Algorithm \ref{alg:dynamic_hedge} line \ref{alg:DH_update1} removes contiguous specialists that are no longer active.
Like GrowingHedge, DynamicHedge can use an unnormalized prior and does not require knowledge of the entire set of experts a-priori.

\section{DynamicHedgeAIXI Agent}
\begin{algorithm}
\caption{DynamicHedgeAIXI}\label{alg:hedge_aixi}
    \begin{algorithmic}[1]
    \State \textbf{Require:} Environment $\mu$, initial history $h_0$, learning rate $\eta > 0$, weights $\bm{\nu} = (\nu_i)_{i \geq 1}$, 
    \State \textbf{Require:} Sequence of sets of contiguous (abstract MDP) specialists $(M_t)_{t \geq 1}$, 
    \State \textbf{Require:} Sequence of composite policies $\pi = (\pi_t)_{t \geq 1}$, where $\pi_t = (\pi_i)_{i \in M_t}$ and $\pi_i: \St_i \to \A$.
    \State \textbf{Initialize:} $L_0 = 0$. For $i \in M_1$, set $w_{1, i} = \nu_i$. 
    \For{$t = 1, 2, \ldots, T$}
        \State Set $\hat{w}_{t, i} = \frac{w_{t, i}}{\sum_{j \in M_t} w_{t, j}}$.
        \State Select $a_t = \arg\max_{a} \sum_{i \in M_t} \hat{w}_{t, i} Q_i^{\pi_i}(h_{<t}, a)$. \label{alg:DHA_action_selection}
        \State Observe $o_t, r_t \sim \mu(\cdot | h_{<t}, a_t)$. 
        \State For all $i \in M_t$, $s^i_t = \phi_i(h_{<t}aor_t)$. 
        \State For all $i \in M_t$, $x_{t, i} = \rho_{t, i}(\cdot| s_{t-1}^{i}, a_t, s_{t}^{i})$ where $\rho_{t, i}: \St_{i} \times \A \times \St_{i} \to \Dist(\Reward)$.
        \State Set $x_t = \frac{\sum_{i \in M_t} w_{t, i} x_{t, i}}{\sum_{i \in M_t} w_{t, i} }$
        \State Set $\ell_t = -\log x_t(r_t)$, $\ell_{t, i} = - \log x_{t, i}(r_t)$.
        \State Set $L_t = L_{t-1} + \ell_t$.
        \State For $i \in M_t \cap M_{t+1}$, set $w_{t+1, i} = w_{t, i} e^{- \eta \ell_{t, i}}$. 
        \State For $i \in M_{t+1} \setminus M_t$, set $w_{t+1, i} = \nu_i e^{- \eta L_t}$.
    \EndFor
    \end{algorithmic}
\end{algorithm}

Our technique for incorporating Dynamic Knowledge Injection, DynamicHedge, can now be naturally integrated into a reinforcement learning agent.
The DynamicHedgeAIXI agent is presented in Algorithm \ref{alg:hedge_aixi}.
We consider the case where specialists are abstract MDPs.
Each specialist $i \in M_t$ produces a function $Q_i^{\pi_i}: \Hist \times \A \to \R$ denoting the expected utility of action $a_t$ under a policy $\pi_i: \St_{i} \to \A$ up to a horizon $H$:
\begin{align}
    Q_{i}^{\pi_i}(h_{<t}, a_t) 
    &= \sum_{sr_{t:t+H}} \left[ \sum_{j=t}^{t+H} r_j \right] \rho_i(sr_{t:t+H} | h_{<t}, a^{i}_{t:t+H}), \label{eqn:specialist_Q}
\end{align}
Also $a^{i}_t = a_t$ and for $k = 1, \ldots, H$ the actions are selected via $a^{i}_{t+k} = \pi_i(s^{i}_{t+k})$.
The agent then selects the action that maximises the weighted sum of the given $Q$ values.

DynamicHedgeAIXI tracks existing and newly entering specialists by computing the weights $w_{t, i}$ using DynamicHedge.
At each time step, specialist $i$ predicts a state-action conditional distribution over the next reward and is evaluated based on the log loss $\ell_{t, i} = - \log \rho_{t, i}(r_t | s_{t-1}^{i}, a_t, s_{t}^{i})$. Thus, over time DynamicHedgeAIXI will weight specialists based on how well they predict the reward sequence over time. 
In this manner, DynamicHedgeAIXI can also avoid the objective mismatch issue common to other system identification approaches to model-based reinforcement learning \cite{lambert2021obj_mismatch,eysenbach2023obj_mismatch}.

\subsection{Properties}
Expanding line \ref{alg:DHA_action_selection}, Algorithm \ref{alg:hedge_aixi} with Equation \ref{eqn:specialist_Q} shows that actions are selected according to:
\begin{multline}
    a_t = \arg\max_{a_t} \\
    \left(\sum_{sr_{t:t+H}} \left[ \sum_{j=t}^{t+H} r_j \right] \sum_{i \in M_t} \hat{w}_{t, i} ~ \rho_i(sr_{t:t+H} | h_{<t}, a^{i}_{t:t+H})\right) \label{eqn:DHA_mix}
\end{multline}
where $\hat{w}_{t, i} = \frac{ w_{t, i} }{ \sum_{j \in M_t} w_{t, j} }$.
Let $\pi_t = (\pi_i)_{i \in M_t}$ denote the composite policy consisting of the policy over each specialist $i \in M_t$. 
DynamicHedgeAIXI's mixture environment model can then be defined as 
\ifedits
\begin{align*}
    \xi^{\pi_t}(r_{t:t+H} | h_{<t}) = \sum_{i \in M_t} \hat{w}_{t, i} ~ \rho_i(r_{t:t+H} | h_{<t}, a^{i}_{t:t+H}),
\end{align*}
\fi
where 
\begin{equation*}
    \begin{split}
    \rho_i(r_{t:t+H} | h_{<t}, a^{i}_{t:t+H}) = \sum_{s_{t:t+H}} \rho_i(sr_{t:t+H} | h_{t}, a^{i}_{t:t+H}).
    \end{split}
\end{equation*}
Expanding $\hat{w}_{t, i} = \frac{w_{t, i}}{\sum_{j \in M_t} w_{t, j}}$ gives
\begin{multline}
    \xi^{\pi_t}(r_{t:t+H} | h_{<t}) = \\
    \sum_{i \in M_t} w_{t}^{i} ~ \rho_{i}(r_{\tau_i:t-1} | sa^{i}_{\tau_i:t-1}) \rho_i(r_{t:t+H} | h_{<t}, a^{i}_{t:t+H}), \label{eqn:DHA_mix_reveal}
\end{multline}
where $w_{t}^{i} = \frac{e^{-L_{\tau_i - 1}}}{ \sum_{j \in M_t} e^{-L_{\tau_j-1}} \rho_{j}(r_{\tau_j:t-1} | sa^{j}_{\tau_j:t-1}) }~$ and $\phi_i(h_{j}) = s^{i}_{j}$ for $j < t$. 

Equation \ref{eqn:DHA_mix_reveal} reveals that $w_{t}^{i}$ can be viewed as the prior weight given to each model in the mixture and that DynamicHedgeAIXI computes an exact Bayesian mixture over the available set of models at each time step.
When a model $i$ first becomes active, its initial weight $w_{\tau_i}^i$ in the mixture environment model depends upon the relative performance of DynamicHedge to each of the available models before time $\tau_i$. 
In this sense, the prior is \textit{adaptive} as it is path dependent.

\subsection{Value Convergence} 
Our main theoretical result shows that DynamicHedgeAIXI will achieve good value convergence rates against the best sequence of environment models available to the agent. 
Since each specialist can potentially operate over a different state space, we first convert the state-reward distribution into a representative observation-reward distribution for our analysis. For a specialist $(\phi, \rho)$, let $\bar{\rho}$ be an environment distribution such that the following hold:
\begin{align*}
    \sum_{\substack{o_t \in \mathcal{\Obs}:\phi(h_{<t}ao_t) = s_t}} \bar{\rho}(o_t | h_{<t}, a_t) &= \rho(s_{t} | s_{t-1}, a_t), \\
    \text{and}\;\;\; \bar{\rho}(r_t | h_{<t}, a_t, o_t) &= \rho(r_t | s_{t-1}, a_t, s_{t}),
\end{align*}
where $\phi(h_{<t}) = s_{t-1}$ and $\phi(hao) = s'$. 
The specialist can then be identified as $(\phi, \bar{\rho})$ and we will drop the bar on $\bar{\rho}$ when the context is clear. 

Most standard Bayesian consistency results compare performance in the realizable setting, where it is assumed that the model class contains the true environment.
To generalise to the dynamic knowledge injection setting where the model class dynamically changes, we instead compare performance against an \textit{admissible} environment sequence. 
\begin{defn}[Admissible environment sequence]
    Let $(M_t)_{t \geq 1}$ be the sequence of sets of specialists. Let $\mu = (\mu_t)_{t \geq 1}$ be an environment sequence such that for all $t \geq 1$ $\mu_t \in M_t$, i.e. there exists $i \in M_t$ where $\rho_i = \mu_t$. We say $\mu$ is admissible if for $j \geq 1$, if $\mu_j \neq \mu_{j+1}$, then $\mu_{j+1} \in M_{j+1} \setminus M_j$.
\end{defn}
Denote by ${V_{i}^H(h_{<t}, \pi_i) = \E^{\pi_i}_{\rho_i} \left[ \sum_{j=t}^{t+H} r_j \vbar h_{<t} \right]}$ the expected future value whilst actions are selected according to $\pi_i$ with specialist $i$. 
Also, let $V_{\xi}^{H}(h_{<t}, \pi_t) = \sum_{i \in M_t} \hat{w}_{t, i} V_i^{H}(h_{<t}, \pi_i)$ denote the expected future value for DynamicHedgeAIXI under the composite policy $\pi_t = (\pi_i)_{i \in M_t}$. 
To simplify analysis, we consider when $\eta = 1$ and $\bm{\nu} = \bm{1}$.

\begin{theorem}
    \label{thm:DHA_value_mu}
    Let $\mu = (\mu_i)_{1 \leq i \leq T}$ be an admissible sequence of environments. 
    Let $\sigma = (\sigma_i)_{1 \leq i \leq k}$ be the switching times such that for all $1 \leq i < k$, $\sigma_i < \sigma_{i+1}$ and for all $1 \leq j \leq T$, $\mu_{j-1} \neq \mu_{j}$ iff $j \in \sigma$. 
    Let $\bar{M}_t = \bigcup_{0 \leq s \leq t} M_s$.
    Then for any sequence of composite policies $\pi = (\pi_t)_{1 \leq t \leq T}$ with $\pi_{\mu_t} \in \pi_t$,
    \begin{multline}
        \sum_{t=1}^{T} \E_{h_{<t} \sim \mu} \left[ \left( V_{\xi}^{H}(h_{<t}, \pi_t) - V_{\mu_{t}}^{H}(h_{<t}, \pi_{\mu_t}) \right)^2 \right] \\
        \leq 2 H^3 r_{\max}^2 \log \frac{1}{w(\mu)}~,
    \end{multline}
    where $w(\mu) = \prod_{j=0}^{k} \hat{w}^{j}_{\tau_j}$ and $\hat{w}^{j}_{\tau_j} = \frac{w_{\tau_j,j}}{\sum_{i \in \bar{M}_T} w_{\tau_j, i}}$. 
\end{theorem}

For a fixed horizon $H$, Theorem \ref{thm:DHA_value_mu} shows that the cumulative squared difference of the value under DynamicHedgeAIXI is bound as a function of $\log \frac{1}{w(\mu)}$. The term $w(\mu)$ can be viewed as the prior weight assigned to the sequence $\mu$ by DynamicHedgeAIXI. In the context of AIXI agents, $w(\mu)$ is a measure of $\mu$'s complexity, with smaller values for $w(\mu)$ indicating a more complex sequence. The complexity of $w(\mu)$ is a function of how often $\mu$ switches and the performance of the models available to the agent. The dependence on the number of switches and available models is given next.

\begin{theorem}
\label{thm:DHA_value_constant}
    Let $\mu = (\mu_i)_{1 \leq i \leq T}$ be an admissible sequence of environments. 
    Let $\sigma = (\sigma_i)_{1 \leq i \leq k}$ be the switching times such that for all $1 \leq i < k$, $\sigma_i < \sigma_{i+1}$ and for all $1 \leq j \leq T$, $\mu_{j-1} \neq \mu_{j}$ iff $j \in \sigma$. 
    Let $\bar{M}_t = \bigcup_{0 \leq s \leq t} M_s$. 
    Then for any sequence of composite policies $\pi = (\pi_t)_{1 \leq t \leq T}$ with $\pi_{\mu_t} \in \pi_t$,
    \begin{multline}
        \sum_{t=1}^{T} \E_{h_{<t} \sim \mu} \left[ \left( V_{\xi}^{H}(h_{<t}, \pi_t) - V_{\mu_{t}}^{H}(h_{<t}, \pi_{\mu_t}) \right)^2 \right] \\ \leq 4 k H^3 r_{\max}^2 \log \abs{\bar{M}_T}~.
    \end{multline}
\end{theorem}

Theorem \ref{thm:DHA_value_constant} makes clear that the cumulative error grows at the rate $\bigO \left( k \log \abs{\bar{M}_T} \right)$. If the number of switches $k$ grows sub-linearly and the total number of models seen by the agent over $T$ steps $\abs{\bar{M}_T}$ grows sub-exponentially in $T$, then DynamicHedgeAIXI's value will converge quickly. 

\subsection{Proof Sketch}
For brevity, we only provide a proof sketch for Theorem \ref{thm:DHA_value_mu}, as Theorem \ref{thm:DHA_value_constant} follows directly from Theorem \ref{thm:DHA_value_mu}.
\ifarxiv
The full proofs of Theorem \ref{thm:DHA_value_mu} and Theorem \ref{thm:DHA_value_constant} are provided in the Appendix.
\fi

Let $\bar{M}_T = \bigcup_{0\leq s \leq T} M_s$ denote the set of all specialists seen by DynamicHedgeAIXI up to time $T$ and let $\pi = (\pi_i)_{i \in \bar{M}_T}$. For each specialist $i \in \bar{M}_T$ we define
\begin{multline*}
    \hat{\rho}^{\pi_i}_i(or_{t:t+H} | h_{<t}) \coloneqq \\
        \begin{cases}
            \; \rho_i^{\pi_i}(or_{t:t+H} | h_{<t}) &\text{if $i \in M_t$}\\
            \; \sum_{j \in M_t} \hat{w}_{t, j} \rho_j^{\pi_j}(or_{t:t+H} | h_{<t}) &\text{if $i \notin M_t$}
        \end{cases}
\end{multline*}
where $\rho_i^{\pi_i}(or_{t:t+H} | h_{<t}) = \rho_i(or_{t:t+H} | h_{<t}, a^{i}_{t:t+H})$ with $a^{i}_{t+k} = \pi_i(s^{i}_{t+k})$ for $k = 0, \ldots, H$.
Using the abstention trick, DynamicHedgeAIXI's value function can then be expressed as follows:
\begin{multline*}
V_{\xi}^{H}(h_{<t}, \pi_t) = \\
\sum_{or_{t:t+H}} \left[ \sum_{j=t}^{t+H} r_j \right] \sum_{j \in \bar{M}_T} \hat{w}_{t, j} \hat{\rho}^{\pi_j}_j(or_{t:t+H} | h_{<t}).
\end{multline*}
We define the mixture model as:
\begin{align*}
    \xi^{\pi}_t(or_{t:t+H} | h_{<t}) = \sum_{j \in \bar{M}_T} \hat{w}_{t, j} \hat{\rho}^{\pi_j}_j(or_{t:t+H} | h_{<t}).
\end{align*}
After expressing the value function in this way, we are ready to bound the error. The sum over $T$ time steps can be first split into a double sum over segments where $\mu$ does not change distribution,
\begin{multline*}
    \sum_{t=1}^{T} \E \left[ \left( V_{\xi}^{H}(h_{<t}, \pi_t) - V_{\mu_{t}}^{H}(h_{<t}, \pi_{\mu_t}) \right)^2 \right] \\
    = \sum_{i=0}^{k} \sum_{t=\sigma_i}^{\sigma_i + 1} \E \left[ \left(V_{\xi}^{H}(h_{<t}, \pi) - V_{\mu_i}^{H}(h_{<t}, \pi_i)\right)^2 \right].
\end{multline*}
We then look to bound the error in each segment. We first convert the squared error between the value functions into a squared error between the mixture distribution and the underlying distribution before applying Pinsker's inequality:
\begin{align*}
    \sum_{t=\sigma_i}^{\sigma_i + 1} &\E \left[ \left( V_{\xi}^{H}(h_{<t}, \pi_t) - V_{\mu_{t}}^{H}(h_{<t}, \pi_{\mu_t}) \right)^2 \right] \\
    &\leq H^2 r_{\max}^2 \sum_{t=\sigma_i}^{\sigma_{i+1} - 1} \E \left[ D_{t:t+H}(\mu_i || \xi^{\pi}_t) \right]\\
    &= H^2 r_{\max}^2 \sum_{t=\sigma_i}^{\sigma_{i+1} - 1} \sum_{n=t}^{t+H} \E \left[ D_{n:n}(\mu_i || \xi^{\pi}_t) \right]
\end{align*}
where $D_{i:j}(\mu || \rho) \coloneqq \sum_{or_{i:j}} \mu(or_{i:j} | h_{<i}) \log \frac{\mu(or_{i:j} | h_{<i})}{\rho(or_{i:j} | h_{<i})}$ denotes the KL divergence on the sequence from time $i$ to $j$. The standard argument is to then apply the chain rule for KL divergence to collapse the double sum into a single KL divergence term.
This cannot be done naively in our case however as the mixture model $\xi^{\pi}_t$ uses a different set of weights at each time step. 
We are however able to apply the chain rule in the standard manner by noticing that the weights that maximize each KL divergence term occur at the start of each segment. This allows us to recover an upper bound of $-\ln \hat{w}_{\sigma_i, i}$ per segment. Summing over all segments gives the final result.

\subsection{DynamicHedgeAIXI in Practice} 
In practice, we let $\eta = 1$, $\bm{\nu} = \bm{1}$ and instantiate DynamicHedgeAIXI in the case where each specialist is a $\Phi$-BCTW model. In this case, each specialist is itself a mixture environment model over a set of $\Phi$-PST environment models. 
Whilst DynamicHedge itself uses an agnostic prior, using $\Phi$-BCTW model's as specialists means that DynamicHedgeAIXI's mixture environment model incorporates a model complexity prior:
\begin{multline}
    \xi^{\pi_t}(r_{t:t+H} | h_{<t}) = \sum_{i \in M_t} \sum_{T \in \mathcal{T}_i} w_{t}^{i} ~ 2^{-\Gamma(T)} \\
    P(r_{\tau_i:t-1} | sa_{\tau_i:t-1}, T) P(r_{t:t+H} | s_{t-1}, a_{t:t+H}, T)~,
\end{multline}
where $\mathcal{T}_i$ is the set of $\Phi$-PST models in the $\Phi$-BCTW specialist $i$.
The $Q$ value functions for each specialist (Algorithm \ref{alg:hedge_aixi}, line 4) also need to be estimated in practice. Each specialist uses the UCT policy \cite{ks06} and Monte-Carlo Tree Search to approximate the finite-horizon expectimax operation in calculating $Q$. 
We show in our experiments that this set of design choices works well in practice.


\section{Experiments}\label{sec:experiments}
\subsection{Experiment Setup}
In the Dynamic Knowledge Injection setting, new knowledge arrives from a human operator in the form of new domain-specific predicates. A predicate environment model is then generated from these predicates for the agent to utilise.
To display the adaptive behaviour of our agent, we consider the setting where better models are generated for the agent over time. 
We simulate the dynamic knowledge injection setting by maintaining two sets of predicates $I$ and $U$ representing the informative and uninformative predicates for each domain. Given a proportion $p \in [0, 1]$, a new $\Phi$-BCTW model of depth $d$ is constructed by sampling $\lfloor p \cdot d \rfloor$ predicates from $I$ and $d - \lfloor p \cdot d \rfloor$ predicates from $U$. The proportion $p$ initially starts out small and increases over time.

In all our experiments, we drop the $\Phi$-BCTW model with the lowest DynamicHedge weight and introduce a new $\Phi$-BCTW model every $4$K steps. The model is also pre-trained on the preceding $4$K steps to ensure it does not perform too poorly to when it is first introduced. The parameter $p$ starts out at $p = 0.05$ and increases by $0.05$ every $4$K steps. 
\ifarxiv
Full details of the experiment design are given in the Appendix.
\fi

\subsubsection{Baseline Methods.} We compare DynamicHedgeAIXI against three baseline methods. We compare against two decision-tree based, iterative state abstraction methods using splitting criteria as defined in U-Tree \cite{McC96} and PARSS-DT \cite{HFD17}. These two models were chosen as they are able to be modified to use predicate functions as state abstraction criteria appropriately. In U-Tree, an existing node (representing a state) is split on a given predicate if splitting results in a statistically significant difference in the resulting Q-values as computed by a Kolmogorov-Smirnov test. In PARSS-DT, nodes are split on a given predicate if the resulting value functions are sufficiently far apart. To modify each method to the dynamic knowledge injection setting, each method can only consider splits from the currently available set of informative predicates as well as the uninformative predicates. This also ensures that the baseline methods do not have an informational advantage over DynamicHedgeAIXI. The third method we compare against is a non-adaptive version of DynamicHedgeAIXI we call HedgeAIXI. HedgeAIXI proceeds in the same way as DynamicHedgeAIXI but does not re-initialize a model's weight when a newly entering model replaces an older model. 

\subsection{Domains}

\subsubsection{Biased Rock-Paper-Scissors (RPS).}
This domain is taken from \cite{FMVRW07}. In biased RPS, the agent plays RPS against an environment with a biased strategy. The environment plays randomly but plays rock if it won the previous round playing rock. There are two informative predicates in this setting: $\it{IsRock}_{t-1}(h)$ and $\it{IsLose}_{t-1}(h)$, indicating whether the environment played rock and whether the agent lost in the previous time step respectively. 

\subsubsection{Taxi.}
The Taxi environment was first introduced in \cite{DI99}. The agent acts as a Taxi in a grid world and must move to pick up a passenger and drop the passenger off at their desired destination. Instead of the 5x5 grid traditionally considered, we consider a 2x5 grid with no intermediate walls and increase the reward for completing the task. These modifications were made to shrink the planning horizon as well as make the original problem less sensitive to parameters for the three algorithms tested.
The predicates available to the agent are indicator functions on the binary representation of the history indicating whether a given bit equals 1. 
The agent can recover the original MDP if it captures the indicator functions  comprising the last received observation.

\subsubsection{Epidemic Control Over Contact Networks.}
The epidemic control problem we use was introduced in \cite{yang-zhao2022a}.
In this environment, an SEIRS epidemic process evolves over a contact network and the agent's goal is to perform actions to slow or stop the spread of the disease \cite{Pastor:2015,Nowzari:2016,Newman:2018}.
A contact network is an undirected graph where the nodes represent individuals and the edges represent interactions between individuals.
We use the network dataset from
\cite{RN15,GDDG03}, which contains 1133 nodes and 5451 edges.
Each node is in one of four states corresponding to their infection status: Susceptible ($\sS$), Exposed ($\sE$), Infectious ($\sI$), and Recovered ($\sR$). Each node also maintains an immunity level.
The environment is partially observable and the environment emits an observation on each node from $\{+, -, ?\}$ corresponding to whether a node tests positive, negative or is unknown/untested. The agent can select actions from a set of 11 possible actions  $\{ \DoNothing, \Vaccinate(i, j), \Quarantine(i) \}$, where $i \in [0, 0.2, 0.4, 0.6, 0.8, 1.0]$ and $j = i+0.2$.
Quarantine actions quarantine the top $i$th percent of nodes ranked by betweenness centrality by removing all edges incident on those nodes for one time step.
Vaccinate actions increase the immunity level (up to a maximum value) for the top $i$th percent of nodes ranked in the same way. 
At each time step, the instantaneous reward is given by
\begin{align*}
    r_t(o_t, a_{t-1}) \coloneqq -Positives(o_t) - Action\_Cost(a_{t-1})
\end{align*}
where $Positives(o_t)$ counts the number of positive tests in the observation $o_t$ and $Action\_Cost(a_{t-1})$ is a function determining the cost of each action. If the agent successfully terminates the epidemic, i.e. there are no more Exposed or Infectious nodes, the agent receives a positive reward of $2$ per node.
\ifarxiv
A full description of the transition and observation models can be found in the Appendix.
\fi

The exact predicates that are considered perfectly useful in this domain are unknown. However, a few are known to provide some information in helping the agent represent the problem. Some examples of the types of functions the agent gains access to over time are as follows:
\begin{itemize}
    \item $\it{ObservedInfectionRate}_{t, \nu}$ takes a history sequence $h \in \Hist$ and computes the observed infection rate at time $t$ over the set of nodes $\nu \subseteq V$. Predicates comparing whether the observed infection rate is greater than or smaller than certain values are received over time.
    \item $\it{InfectionRateOfChange}_{t, \nu}$ takes $h \in \Hist$ and computes the change in infection rate between timesteps $t-1$ and $t$ over the set of nodes $\nu \subseteq V$. Predicates comparing whether the infection rate of change is greater than or smaller than certain values are received over time.
    \item $\it{PercentAction}_{a, N}$ takes a history and returns the percentage of time action $a$ was selected in the last $N$ timesteps. Predicates comparing whether the action selection percentage is greater than or smaller than certain values are received over time. 
\end{itemize}

Epidemic control is a topical subject given the prevalance of COVID-19 and recent approaches to this question vary wildly in terms of the how the problem is modelled \cite{ArPe20,CELT20,CHRTOMP20,Brauer:2012,Anderson:2013,BKSE21}.
With no consensus on the appropriate model to use, our model is chosen as it is sufficiently complex to demonstrate the efficacy of our agent. 

\subsection{Results}

\begin{figure}[t]
    \centering
    \resizebox{0.95\columnwidth}{!}{\begin{tabular}{c}
        \includegraphics{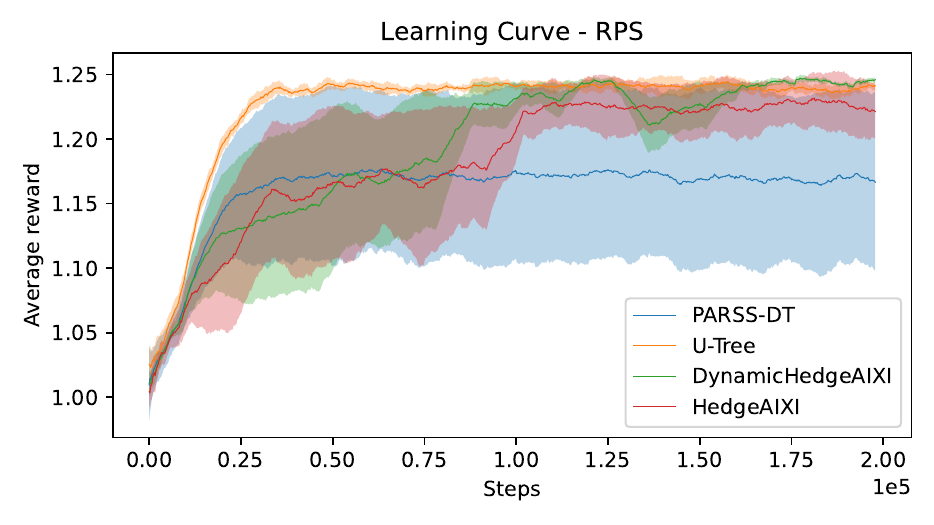} \\
        \includegraphics{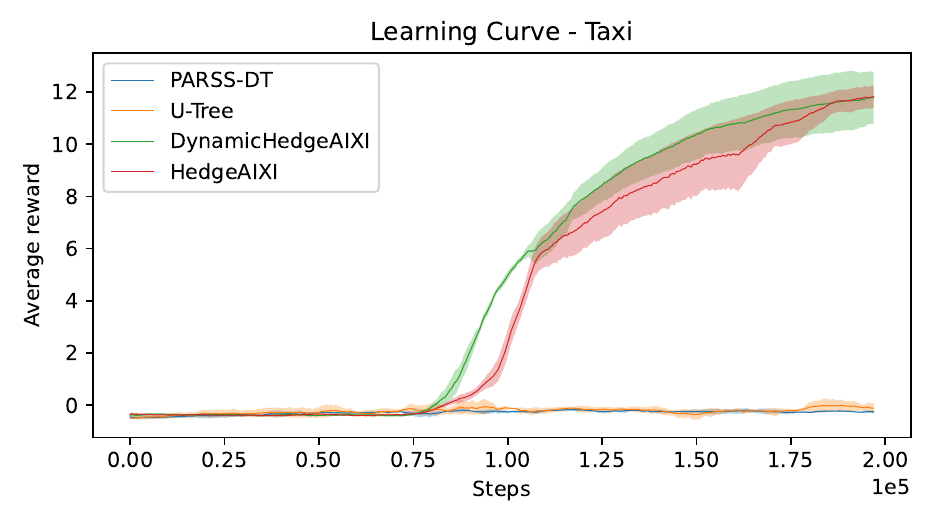}
    \end{tabular}}
    \caption{Learning curve on biased RPS and Taxi domains.}
    \label{fig:rps_taxi}
\end{figure}

\begin{figure}[t]
    \centering
    \resizebox{0.95\columnwidth}{!}{\begin{tabular}{c}
        \includegraphics{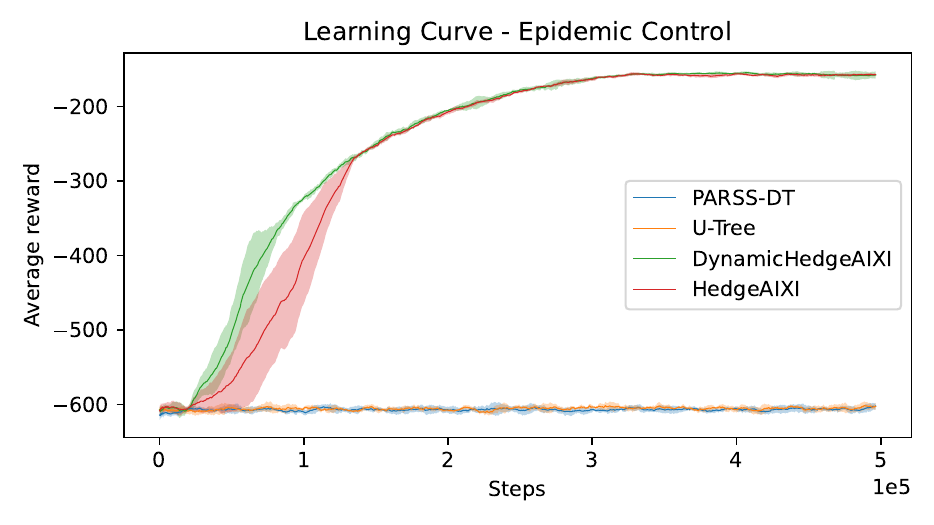} \\
        \includegraphics[scale=1.27]{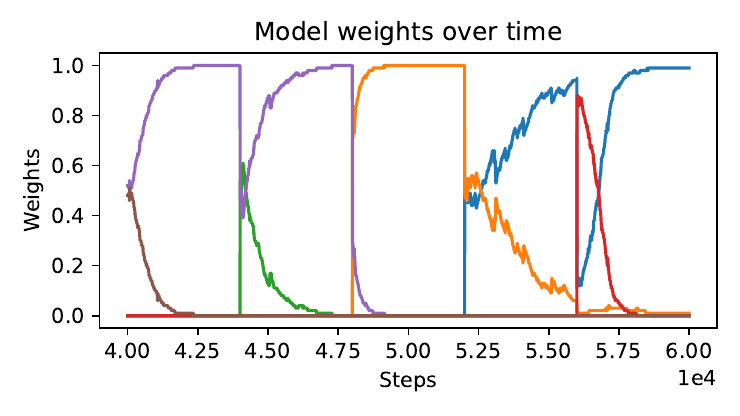}
    \end{tabular}}
    \caption{Epidemic control: plot of learning curve and agent's model weights over time.}
    \label{fig:JP_EC}
\end{figure}

Figures \ref{fig:rps_taxi} and \ref{fig:JP_EC} display the mean learning curves for DynamicHedgeAIXI, U-Tree, 
PARSS and HedgeAIXI with standard deviations computed over five random seeds. DynamicHedgeAIXI and HedgeAIXI vastly outperform U-Tree and PARSS on the Taxi and Epidemic Control domains. 
PARSS and U-Tree were likely unable to perform because these two domains require a large number of predicates to be 
evaluated together rather than individually for node splits. Also, both methods can be susceptible to producing spurious splits, making the resulting state space too difficult to learn with.
On both Taxi and Epidemic Control, the learning curves for both DynamicHedgeAIXI and HedgeAIXI start out flat due to the lack of 
useful environment models initially. As useful models are introduced, both agents' performance begin to pick up. 
DynamicHedgeAIXI's performance picks up faster than HedgeAIXI. This effect is especially pronounced in the Taxi domain. 
HedgeAIXI's slower convergence on these more complex domains is likely due to it taking longer to overcome the bad weights 
left behind by previous models. This highlights the fast adaptation that DynamicHedge can provide.
On the RPS environment, DynamicHedgeAIXI and HedgeAIXI were more unstable compared to U-Tree.
DynamicHedgeAIXI's initial instability likely comes from the frequent re-distribution of posterior weight towards newly entering models. 
In contrast, since the environment is rather small and requires only two predicates to fully model, U-Tree was able to find the correct splits quickly and did not consider additional splits.
Nevertheless, DynamicHedgeAIXI eventually converges to optimal performance.

To further demonstrate DynamicHedgeAIXI's adaptive behaviour, Figure \ref{fig:JP_EC} displays the model weights over time during the 40k to 60k steps period of a single run of the Epidemic Control problem. The sharp drops in weight for the highest weighted model correspond to when a new model enters. Similarly, the sharp spikes in a coloured line indicate when a previous model has been removed and replaced with a new model. Between models being introduced, the weights converge quickly to the best performing model. Over this period of time, the highest weighted model switches to a newly entering model twice. 
Considering the algorithm's performance in the Epidemic Control domain, DynamicHedgeAIXI can be seen as adapting effectively. 

\section{Discussion and Conclusion}

The key contribution of this paper is the introduction of an exact and efficient Bayesian mixture modelling algorithm for general reinforcement learning agents in the dynamic knowledge injection setting, a setting that formalises a form of Human-AI teaming construct. The algorithm generalizes and integrates Hedge and CTW, two highly successful Bayesian mixture models, and we provide theoretical and empirical results that demonstrate the algorithm's practical utility. Our research opens the door to the principled construction of collaborative Human-AI teaming systems for complex environments, where (a) the human expert can use an expressive knowledge representation formalism to supply the agent with new background knowledge in an interactive manner to reduce model bias in the agent's initial knowledge base, and (b) the AI agent can learn to act optimally with respect to an exact Bayesian mixture model that comes with performance guarantees measured against the best aspects of human knowledge injection. Our work is another proof point that the AIXI theory can provide good guidance on the design of new general reinforcement learning agents. A key limitation of our approach is the transfer of a huge burden of learning to the human expert. This can be alleviated by augmenting the human expert with a semi-automatic predicate generation process. The predicate-selection algorithm described in \cite{yang-zhao2022a} can be run in micro-batches to instantiate that external process. More generally, the use of statistical predicate invention \cite{CropperMM20,KokD07} techniques is an important direction for future work.

\newpage

\section{Acknowledgments}
The authors would like to thank Joel Veness, Elliot Catt, Jordi Grau-Moya, and Tim Genewein for their thoughtful feedback on an earlier draft. This work was in parts supported by ARC grant DP150104590.

\bibliography{aaai24}

\ifarxiv
\appendix
\newpage
\section{Appendix}

\section{Theoretical Results}
\label{appendix:theory}

\subsection{Proof of Theorem \ref{thm:DHA_value_mu}}

\begin{T1}
    Let $\mu = (\mu_i)_{1 \leq i \leq T}$ be an admissible sequence of environments. 
    Let $\sigma = (\sigma_i)_{1 \leq i \leq k}$ be the switching times such that for all $1 \leq i < k$, $\sigma_i < \sigma_{i+1}$ and for all $1 \leq j \leq T$, $\mu_{j-1} \neq \mu_{j}$ iff $j \in \sigma$. 
    Then for any sequence of composite policies $\pi = (\pi_t)_{1 \leq t \leq T}$ with $\pi_{\mu_t} \in \pi_t$,
    \begin{multline}
        \sum_{t=1}^{T} \E_{h_{<t} \sim \mu} \left[ \left( V_{\xi}^{H}(h_{<t}, \pi_t) - V_{\mu_{t}}^{H}(h_{<t}, \pi_{\mu_t}) \right)^2 \right] \\
        \leq 2 H^3 r_{\max}^2 \log \frac{1}{w(\mu)}~,
    \end{multline}
    where $w(\mu) = \prod_{j=0}^{k} w_{\tau_j}^j$. 
\end{T1}

\begin{proof}
    The proof is split into two parts. We first define some notation and properties we will need before proving the theorem.

    \noindent\textbf{Part 1: Properties}
    
    \noindent \textbf{Induced environment models and policies.}
    For an abstract MDP $(\phi, \rho)$ where $\rho: \St \times \A \to \Dist(\St \times \Reward)$, let $\Obs(h_{<t}, sa_t) = \{o_t \in \Obs : \phi(h_{<t}ao_t) = s_t\}$.We can then define $\bar{\rho}(or_t | h_{<t}, a_t) = \bar{\rho}(r_t | h_{<t}, a_t, o_t) \bar{\rho}(o_t | h_{<t}, a_t)$ such that 
    $\sum_{o_t \in \Obs(h_{<t}, sa_t)} \bar{\rho}(o_t | h_{<t}, a_t) = \rho(s_t | s_{t-1}, a_t)$ and $\bar{\rho}(r_t | h_{<t}, a_t, o_t) = \rho(r_t | s_{t-1}, a_t)$ where $\phi(h_{<t}) = s_{t-1}$ and $\phi(h_{<t}ao_t) = s_t$. 
    For example, we can define $\bar{\rho}(o_{t} | h_{<t}, a_t)$ as:
    \begin{align*}
        \bar{\rho}(o_t | h_{<t}, a_t) = \frac{1}{\abs{\Obs(h_{<t}, sa_t)}} \rho(s_t | s_{t-1}, a_t).
    \end{align*}
    Also, for a state-dependent policy $\pi: \St \to \A$ we can define the induced history-dependent policy as $\bar{\pi}(h_{<t}) = \pi(\phi(h_{<t}))$. The crucial property is that the following equivalence holds:
    \begin{multline}
        \sum_{o_t \in \Obs} \rho(or_t | h_{<t}, \bar{\pi}(h_{<t}))
        = \sum_{s_t \in \St} \rho\left(sr_t | s_{t-1}, \pi(s_{t-1})\right) \label{eqn:1}
    \end{multline}
    When the context is clear, we drop the superscript on $\bar{\rho}$ and $\bar{\pi}$. Also let $\rho(sr_t | h_{<t}, a_t) = \rho(sr_t | s_{t-1}, a_t)$ where $\phi(h_{<t}) = s_{t-1}$.

    Recall that the value function under $\rho$ and $\pi$ can be expressed as:
    \begin{align*}
        V^{H}_{\rho}(h_{<t}, \pi) = \sum_{sr_{t:t+H}} \left[ \sum_{j=t}^{t+H} r_j \right] \rho(sr_{t:t+H} | h_{<t}, a_{t:t+H}),
    \end{align*}
    where $a_k = \pi(s_{k-1})$ for $k = t, \ldots, t+H$. 
    Using Equation \ref{eqn:1}, we can also express the value function in terms of observation-rewards:
    \begin{align}
        V_{\rho}^{H}(h_{<t}, \pi) &= \sum_{sr_{t:t+H}} \left[ \sum_{j=t}^{t+H} r_j \right] \rho(sr_{t:t+H} | h_{<t}, a_{t:t+H}) \nonumber\\
        &= \sum_{or_{t:t+H}} \left[ \sum_{j=t}^{t+H} r_j \right] \rho(or_{t:t+H} | h_{<t}, a_{t:t+H}). \label{eqn:2}
    \end{align}
    The actions in Equation \ref{eqn:2} are well-defined under the history-dependent policy $\bar{\pi}$.
    For notational convenience, let $\rho^{\pi}(or_{t:t+H} | h_{<t}) = \rho(or_{t:t+H} | h_{<t}, a_{t:t+H})$. 
 
    We can now also express DynamicHedgeAIXI's value function. Define the mixture environment model as:
    \begin{align*}
        \xi_j^{\pi_j}(or_{t:t+H} | h_{<t}) = \sum_{i \in M_j} \frac{w_{j, i}}{\sum_{i \in M_j} w_{j, i}} \rho_i^{\pi_i}(or_{t:t+H} | h_{<t}).
    \end{align*}
    The subscript $j$ indicates which time step the models in the mixture are from and also which time step the weights are from.
    
    DynamicHedgeAIXI's value function $V^{H}_{\xi}$ under $\pi_t$ can be expressed using the mixture environment model as follows:
    \begin{multline*}
        V_{\xi}^{H}(h_{<t}, \pi_t) = \sum_{i \in M_t} \frac{w_{t, i}}{\sum_{i \in M_t} w_{t, i}} V_{\rho_i}^{H}(h_{<t}, \pi_i)\\
        = \sum_{i \in M_t} \frac{w_{t, i}}{\sum_{i \in M_t} w_{t, i}} \sum_{or_{t:t+H}} \left[ \sum_{j=t}^{t+H} r_j \right] \rho_i^{\pi_i}(or_{t:t+H} | h_{<t})\\
        = \sum_{or_{t:t+H}} \left[ \sum_{j=t}^{t+H} r_j \right] \sum_{i \in M_t} \frac{w_{t, i}}{\sum_{i \in M_t} w_{t, i}} \rho_i^{\pi_i}(or_{t:t+H} | h_{<t})\\
        = \sum_{or_{t:t+H}} \left[ \sum_{j=t}^{t+H} r_j \right] \xi_t^{\pi_t}(or_{t:t+H} | h_{<t}).
    \end{multline*}

    \textbf{Predictive distribution.}
    For any environment model $\rho$ (over observation-rewards), the distribution over the next observation-reward is given by:
    \begin{align*}
        \rho^{\pi}(or_t | h_{<t}) = \frac{ \rho^{\pi}(or_{1:t}) }{ \rho^{\pi}(or_{1:t-1}) }.
    \end{align*}
    This implies for $i \leq j$, $\rho^{\pi}(or_{i:j} | h_{<i}) = \frac{ \rho^{\pi}(or_{1:j}) }{ \rho^{\pi}(or_{1:i-1}) }$. 

    \textbf{Mixture model and the abstention trick.}  
    We can rewrite the mixture model using the abstention trick to instead be a mixture over all models $i \in \bar{M}_T$. 
    For an environment model $\rho$, define $\hat{\rho}$ as:
    \begin{multline*}
        \hat{\rho}(or_{t:t+H} | h_{<t}, a_t) \\
        = 
            \begin{cases}
                \rho(or_{t:t+H} | h_{<t}, a_t) &\text{if $i \in M_t$}\\
                \sum_{i \in M_t} \frac{w_{t, i}}{\sum_{i \in M_t} w_{t, i}} \rho_i(or_{t:t+H} | h_{<t}, a_t) &\text{if $i \notin M_t$}
            \end{cases}
    \end{multline*}
    Recall that the weights are defined as $w_{t, i} = e^{-L_{t-1, i}}$ (since the prior weight is equal to 1) where $L_{t-1, i} = \sum_{s \leq t-1: i \in M_s} \ell_{s, i} + \sum_{s \leq t-1: i \notin M_s} \ell_{s, i}$. 
    Let $\bar{M}_T = \bigcup_{1 \leq s \leq T} M_{s}$ denote the set of all models seen over $T$ steps and let $\pi = (\pi_i)_{i \in \bar{M}_T}$.
    Using $\hat{\rho}$, the mixture model $\xi^{\pi_t}_t$ is given by:
    \begin{multline}
        \sum_{i \in M_t} \frac{w_{t, i}}{\sum_{i \in M_t} w_{t, i}} \rho_i^{\pi_i}(or_{t:t+H} | h_{<t}) \\
        = \sum_{i \in \bar{M}_T} \frac{w_{t, i}}{\sum_{i \in \bar{M}_T} w_{t, i}} \hat{\rho}_{i}^{\pi_i}(or_{t:t+H} | h_{<t}) \\
        = \sum_{i \in \bar{M}_T} \hat{w}_{j, i} \hat{\rho}_i^{\pi_i}(or_{t:t+H} | h_{<t}), \label{eqn:mix_1}
    \end{multline}
    where $\hat{w}_{j, i} = \frac{w_{j, i}}{\sum_{i \in \bar{M}_T} w_{j, i}}$. The subscript $j$ now indicates which time step the weights are from. 
    Let $\xi_j^{\pi}(or_{t:t+H} | h_{<t}) = \sum_{i \in \bar{M}_T} \hat{w}_{j, i} \hat{\rho}_i^{\pi_i}(or_{t:t+H} | h_{<t})$. 
    From Equation \ref{eqn:mix_1}, DynamicHedgeAIXI's value function can now be expressed as:
    \begin{align*}
        V_{\xi}^{H}(h_{<t}, \pi_t) &= V_{\xi}^{H}(h_{<t}, \pi)\\
        &= \sum_{or_{t:t+H}} \left[ \sum_{j=t}^{t+H} r_j \right] \xi^{\pi}_t(or_{t:t+H} | h_{<t}).
    \end{align*}

    We will compare the value error when the value function is expressed using $\hat{\rho}$. 
    \hfill \newline

    \textbf{Part 2: Bounding the error}

    To bound the error, we first split the single sum over $t$ into a double sum over each segment between switches in the sequence $\mu$. Let $\sigma_0 = 0$ and $\sigma_{k+1} = T+1$. 
    Since $\mu_m = \mu_n$ for $\sigma_i \leq m, n \leq \sigma_{i+1}-1$, let $i$ denote the model between times $\sigma_i$ and $\sigma_{i+1}-1$ and let $\bar{\mu}_i$ denote its distribution. We thus have
    \begin{multline}
        \sum_{t = 1}^{T} \E_{h_{<t} \sim \mu} \left[ \left( V_{\xi}^{H}(h_{<t}, \pi) - V_{\mu_t}^{H}(h_{<t}, \pi_{\mu_t}) \right)^2 \right] \\
        \leq 
        \sum_{i = 0}^{k} \E\left[ \sum_{t=\sigma_i}^{\sigma_{i+1}-1} \E\left[ \left( V_{\xi}^{H}(h_{<t}, \pi) - V_{i}^{H}(h_{<t}, \pi_{i}) \right)^2 \right] \right]. \label{eqn:0}
    \end{multline}
    For $i \leq j$, define the KL divergence between two distributions over an observation-reward sequence as:
    \begin{align*}
        D_{i:j}(\mu^{\pi} || \rho^{\bar{\pi}}) = \sum_{or_{i:j}} \mu^{\pi}(or_{i:j} | h_{<i}) \log \frac{\mu^{\pi}(or_{i:j} | h_{<i})}{\rho^{\bar{\pi}}(or_{i:j} | h_{<i})}.
    \end{align*}

    Using Pinsker's inequality and the chain rule for KL divergence, the squared error can be bound as follows:
    \begin{equation}
    \begin{split}
        &\left( V_{\xi}^{H}(h_{<t}, \pi) - V_{\bar{\mu}_i}^{H}(h_{<t}, \pi_i) \right)^2 \\
        &\leq \left( \sum_{or_{t:t+H}} \left[ \sum_{j=t}^{t+H} r_j \right] \left(\xi_t^{\pi} - \bar{\mu}_i^{\pi_i}\right) \right)^2 \\
        &\leq 2 H^2 r_{\max}^2 D_{t:t+H}(\bar{\mu}_i^{\pi_i} || \xi^{\pi}_t) \\
        &= 2 H^2 r_{\max}^2 \sum_{n = t}^{t+H} \E_{h_{t:n-1} \sim \bar{\mu}_i} \left[ D_{n:n}(\bar{\mu}_i || \xi^{\pi}_t) \right].  \label{eqn:3}
    \end{split}
    \end{equation}

    Using Equation \ref{eqn:3}, over a single segment we have:
    \begin{multline}
        \sum_{t=\sigma_i}^{\sigma_{i+1}-1} \E_{h_{\sigma_i:t-1} \sim \bar{\mu}_i} \left[ \left( V_{\xi}^{H}(h_{<t}, \pi) - V_{\bar{\mu}_i}^{H}(h_{<t}, \pi_i) \right)^2 \right] \\
        \leq 
        2 H^2 r_{\max}^2 \sum_{t=\sigma_i}^{\sigma_{i+1}-1} \sum_{n=t}^{t+H} \E_{h_{\sigma_i:n-1} \sim \bar{\mu}_i} \left[ D_{n:n}(\bar{\mu}_i || \xi_n^{\pi}) \right]. \label{eqn:4}
    \end{multline}

    Let $G = \sigma_{i+1}-1 + H - \sigma_i$ and for $\sigma_i \leq t \leq \sigma_i + G$, let $l = \arg\max_{\sigma_i \leq n \leq \sigma_i + G} \left[ D_{t:t}(\bar{\mu}_i || \xi_{n}^{\pi}) \right]$. 
    Note that for each $t = \sigma_i, \ldots, \sigma_i+G$, the KL divergence term $D_{t:t}(\cdot || \cdot)$ appears at most $H$ times. Thus we have
    \begin{equation}
    \begin{split}
        &\sum_{t = \sigma_i}^{\sigma_{i+1}-1} \sum_{n = t}^{t+H} \E_{h_{\sigma_i:n-1} \sim \bar{\mu}_i} \left[ D_{n:n}(\bar{\mu}_i || \xi_n^{\pi}) \right] \\
        &\leq H \sum_{t=\sigma_i}^{\sigma_{i}+G} \E_{h_{\sigma_i:t-1} \sim \bar{\mu}_i} \left[ D_{t:t}(\bar{\mu}_i || \xi_{l}^{\pi}) \right]\\
        &= H \cdot D_{\sigma_i: \sigma_{i+G}} (\bar{\mu}_i || \xi^{\pi}_{l}).
    \end{split}
    \end{equation}

    Using the fact that $\bar{\mu}_i \in \bar{M}_T$, the KL divergence term can be bound as follows:
    \begin{align}
        D_{\sigma_i: \sigma_{i+G}} (\bar{\mu}_i || \xi^{\pi}_{l}) &= \E_{h_{\sigma_{i}:\sigma_{i}+G} \sim \bar{\mu}_i} \left[ \ln \frac{\bar{\mu}_i(or_{\sigma_i:\sigma_{i}+G} | h_{<\sigma_i})}{\xi^{\pi}_l(or_{\sigma_i:\sigma_{i}+G} | h_{<\sigma_i})} \right] \nonumber \\
        &\leq \E_{h_{\sigma_{i}:\sigma_{i}+G} \sim \bar{\mu}_i} \left[ \ln \frac{1}{\hat{w}_{l, i}} \right]. \label{eqn:mix_prior_bound}
    \end{align}
    Define $L_{\sigma_i, j}^{l} = L_{l, j} - L_{\sigma_i, j}$. 
    Let $m = \arg\min_{j \in \bar{M}_T} L_{\sigma_i, j}^{l}$. 
    Then we have:
    \begin{equation}
    \begin{split}
        &\E_{h_{\sigma_{i}:\sigma_{i}+G} \sim \bar{\mu}_i} \left[ \ln \frac{1}{\hat{w}_{l, i}} \right] \\
        &= \E_{h_{\sigma_{i}:\sigma_{i}+G} \sim \bar{\mu}_i} \left[ \ln \frac{\sum_{j \in \bar{M}_T} w_{l, j} }{w_{l, i}} \right]\\
        &= \E_{h_{\sigma_{i}:\sigma_{i}+G} \sim \bar{\mu}_i} \left[ \ln \frac{\sum_{j \in \bar{M}_T} w_{\sigma_i, j} e^{-L_{\sigma_i, j}^{l}} }{w_{\sigma_i, i} e^{-L_{\sigma_i, i}^{l}} } \right]\\
        &\leq \log \frac{1}{\hat{w}_{\sigma_i, i}} + \E_{h_{\sigma_{i}:\sigma_{i}+G} \sim \bar{\mu}_i} \left[ - L_{\sigma_i, m}^{l} + L_{\sigma_i, i}^{l}\right].
    \end{split}
    \end{equation}
    The inequality arises by upper bounding each $e^{-L_{\sigma_i,j}^{l}}$ term in the numerator with $e^{-L_{\sigma_i,m}^{l}}$. 
    
    The term $\E_{h_{\sigma_{i}:\sigma_{i}+G} \sim \bar{\mu}_i} \left[ - L_{\sigma_i, m}^{l} + L_{\sigma_i, i}^{l}\right]$ should be upper bound by 0 since model $i$ should be the model with the smallest loss in expectation. We show this formally.
    
    Recall for a model $j$, the loss at time $t$ is given by $\ell_{t, j} = - \ln \rho_j(r_t | s_{t-1}, a_t, s_t)$ and we have $\rho_j(r_t | s_{t-1}, a_t, s_t) = \rho_j(r_t | h_{<t}, a_t, o_t)$. Thus,
    \begin{align*}
        L_{\sigma_i, j}^{l} &= - \sum_{t = \sigma_i}^{l} \ell_{t, i} \\
        &= - \sum_{t = \sigma_i}^{l} \ln \rho_j(r_t | h_{<t}, a_t, o_t).
    \end{align*}
    Thus the expectation is bound as follows:
    \begin{equation}
    \begin{split}
        &\E \left[ - L_{\sigma_i, m}^{l} + L_{\sigma_i, i}^{l}\right] \\
        &= \sum_{t = \sigma_i}^{l} \E\left[ \sum_{r_t} \bar{\mu}_i^{\pi_i}(r_t | h_{<t}, a_t, o_t) \ln \frac{ \rho^{\pi_m}_{m}(r_t | h_{<t}, a_t, o_t) }{ \bar{\mu}_i^{\pi_i}(r_t | h_{<t}, a_t, o_t) } \right]\\
        &\leq 0.
    \end{split}
    \end{equation}
    The last inequality follows since term inside the square bracket is the negative KL divergence, which is bound below 0. 

    Thus, substituting terms back into Equation \ref{eqn:4}, the bound over a single segment is given by:
    \begin{equation}
    \begin{split}
        &\sum_{t=\sigma_i}^{\sigma_{i+1}-1} \E \left[ \left( V_{\xi}^{H}(h_{<t}, \pi) - V_{\bar{\mu}_i}^{H}(h_{<t}, \pi_i) \right)^2 \right] \\
        &\leq 2 H^3 r_{\max}^2 \log \frac{1}{\hat{w}_{\sigma_i, i}} \\
        &= 2 H^3 r_{\max}^2 \log \frac{1}{\hat{w}_{\tau_i}^{i}} \label{eqn:6}
    \end{split}
    \end{equation}
    Combining Equation \ref{eqn:6} over $k$ segments, substituting back into Equation \ref{eqn:0}, and applying the definition of $w(\mu)$ gives the final result.    
\end{proof}

\subsection{Proof of Theorem \ref{thm:DHA_value_constant}}

Theorem \ref{thm:DHA_value_constant}, restated below, follows from Theorem \ref{thm:DHA_value_mu} by bounding $w(\mu)$.

\begin{T2}
    Let $\mu = (\mu_i)_{1 \leq i \leq T}$ be an admissible sequence of environments. Let $\sigma = (\sigma_i)_{1 \leq i \leq k}$ be the switching times such that $\sigma_i < \sigma_{i+1}$ and $\mu_{i-1} \neq \mu_{i}$ only when $i = \sigma_{i}$.
    Let $\bar{M}_t = \bigcup_{0 \leq s \leq t} M_s$. 
    Then for any sequence of composite policies $\pi = (\pi_t)_{1 \leq t \leq T}$ with $\pi_{\mu_t} \in \pi_t$,
    \begin{align}
        \sum_{t=1}^{T} \E_{h_{<t} \sim \mu} \left[ \left( V_{\xi}^{H}(h_{<t}, \pi_t) - V_{\mu_{t}}^{H}(h_{<t}, \pi_{\mu_{t}}) \right)^2 \right] \nonumber\\
        \leq 4 k H^3 r_{\max}^2 \log \abs{\bar{M}_T}~.
    \end{align}
\end{T2}

\begin{proof}
    Start with the bound provided in Theorem \ref{thm:DHA_value_mu}. The term $w(\mu)$ is given by ${w(\mu) = \sum_{j=1}^{k} \log \frac{1}{\hat{w}_{\tau_j}^j}}$. The weight $w_{\tau_j}^j$ can be expressed as
    \begin{align*}
        \hat{w}_{\tau_j}^j &= \frac{ e^{-L_{\tau_j-1, j}} }{ \sum_{i \in \bar{M}_{T}} e^{-L_{\tau_j-1, i}} }\\
        &= \frac{ 1 }{ \sum_{i \in \bar{M}_{T}} e^{L_{\tau_j-1, j}-L_{\tau_j-1, i}} }
    \end{align*}

    Recall that $L_{t, i} = \sum_{s \leq t: i \in M_s} \ell_{s, i} + \sum_{s \leq t: i \notin M_s} \ell_s$. So each model incurs DynamicHedge's loss until the time they enter. Therefore only models $i \in \bar{M}_T$ with $\tau_i < \tau_j$ differ in their loss compared to model $j$. Thus for any $i \in \bar{M}_T$ such that $\tau_i < \tau_j$,
    \begin{align*}
        L_{\tau_j - 1, j} - L_{\tau_j - 1, i} &= \sum_{t = \tau_i}^{\tau_j - 1} \ell_t - \ell_{t,i}\\
        &\leq \log \abs{\bar{M}_T}~,
    \end{align*}
    where $\bar{M}_T = \bigcup_{1 \leq s \leq t} M_s$ is the set of all models seen by the agent up to time $T$. The inequality follows from Hedge's regret bound (Proposition \ref{prop:hedge_regret}) and generating a prior from the unnormalized prior weight by multiplying by $\frac{1}{\abs{\bar{M}_T}}$. Thus we have that $\hat{w}_{\tau_j}^j$ is lower bounded by
    \begin{align*}
        w_{\tau_j}^j &\geq \frac{1}{\sum_{i \in \bar{M}_T} \abs{\bar{M}_T}}\\
        &\geq \frac{1}{\abs{\bar{M}_T}^2}.
    \end{align*}
    Substituting this back into our value bound gives our final result.
\end{proof}

\section{Design of Experiments}
\label{appendix:design_of_experiments}
We describe the details of our experiment design in this section. 
All experiments were performed on a shared server with a 32-Core Intel(R) Xeon(R) Gold 5218 CPU and 192 gigabytes of RAM.

\subsection{Dynamic Knowledge Injection Setup}
We simulate the dynamic knowledge injection setting by maintaining two sets of predicates $I$ and $U$ representing the informative and uninformative predicates for each domain. Given a proportion $p \in [0, 1]$, a new $\Phi$-BCTW model of depth $d$ is constructed by sampling $\lfloor p \cdot d \rfloor$ predicates from $I$ and $d - \lfloor p \cdot d \rfloor$ predicates from $U$. The proportion $p$ initially starts out small and increases over time.

In all our experiments, we drop the $\Phi$-BCTW model with the lowest DynamicHedge weight and introduce a new $\Phi$-BCTW model every $4$K steps. The model is also pre-trained on the preceding $4$K steps to ensure it does not perform too poorly to when it is first introduced. The parameter $p$ starts out at $p = 0.05$ and increases by $0.05$ every $4$K steps. 

\subsection{Environment Design}
We now describe the design of each environment, the parameters used, and the predicates used.
The following functions help us deal with binary representations:
\begin{itemize}
    \item $\it{Encode}_i$ splits the possible range of its argument into $2^i$ equal sized buckets and encodes its argument by the number of whichever bucket it falls into. If the range is unbounded, it is first truncated. 
    \item $\it{Bit}_i$ takes a bit string and returns the $i$th bit. 
    \item $(= 1)$ returns whether the provided argument equals 1.
    \item $(\geq p)$ returns whether the provided argument is greater than $p$.
\end{itemize}

Function composition\index{composition ($\comp$)} is handled by the (reverse)
composition function 
\[ \comp : (a \rightarrow b) \rightarrow (b \rightarrow c) \rightarrow 
             (a \rightarrow c) \]
defined by
$ ((f \comp g) \; x) = (g \; (f \; x)).$

\textbf{Biased Rock-Paper-Scissors (RPS).} 
As mentioned in Section 5.1, there are two informative predicates in this domain: $IsRock_{t-1}(h)$ and $IsLose_{t-1}(h)$ indicate whether the envionment played rock and whether the agent lost in the previous time step respectively. The $\Phi$-BCTW models introduced in this domain are all constructed to have depth 2. 

\textbf{Taxi.} 
We modify the original taxi environment of \cite{FMVRW07}, such that a 2x5 grid is used instead. The passenger and destination still spawn in the corner locations with coordinates $(0,0), (1,0), (0, 4), (1,4)$. The agent receives -1 reward for running into a wall, +100 reward for successfully dropping off the passenger at its destination. The observation received at each time step is the state $(x, y, p, d)$ where $(x, y)$ denote the position of the agent (representing the taxi), $p$ is an index indicating the passenger position, and $d$ is an index indicating the destination position. The agent has access to informative predicate functions of the following forms:
\begin{itemize}
    \item $\it{xDistToPassenger}_{1} \comp Encode_n \comp Bit_i \comp (= 1)(h)$ - $\it{xDistToPassenger}_{1}$ takes a history sequence and returns the $x$ distance from the agent to the passenger at the last time step. The predicate returns whether the $i$th bit of the $x$ distance encoded into $n$ bits equals 1. 
    \item $\it{yDistToPassenger}_{1} \comp Encode_n \comp Bit_i \comp (= 1)(h)$ - $\it{yDistToPassenger}_{1}$ takes a history sequence and returns the $y$ distance from the agent to the passenger at the last time step. The predicate returns whether the $i$th bit of the $y$ distance encoded into $n$ bits equals 1. 
    \item $\it{xDistToDestination}_{1} \comp Encode_n \comp Bit_i \comp (= 1)(h)$ - $\it{xDistToDestination}_{1}$ takes a history sequence and returns the $x$ distance from the agent to the destination at the last time step. The predicate returns whether the $i$th bit of the $x$ distance encoded into $n$ bits equals 1. 
    \item $\it{yDistToDestination}_{1} \comp Encode_n \comp Bit_i \comp (= 1)(h)$ - $\it{yDistToDestination}_{1}$ takes a history sequence and returns the $y$ distance from the agent to the destination at the last time step. The predicate returns whether the $i$th bit of the $y$ distance encoded into $n$ bits equals 1. 
    \item $\it{PassengerPickedUp}(h)$ - Takes a history sequence $h$ and returns whether the passenger was in the taxi at the last time step. 
    \item $\it{Suffix}_N \comp (= 1) (h)$ - Returns whether the $N$th bit of the suffix of the history in binary representation equals 1. The agent will recover the original MDP if it captures the indicator functions that comprise the last received observation.
\end{itemize}

The $\Phi$-BCTW models introduced in this domain are constructed from a set of $17$ predicates. 

\textbf{Epidemic Control over Contact Networks.}
We delay the full description of the epidemic control environment and the parameters used but describe the predicates used here. We assume that the agent has a background theory consisting of:
\begin{itemize}
    \item A graph $G = (V, \mathcal{E})$ that captures (approximately) the structure of the underlying contact network, but ont the disease status of nodes. The connectivity between individuals could be inferred from census data, telecommunication records, contact tracing apps etc.
    \item The transition and observation functions of a dynamics model of the underlying disease but not the parameters. This information would be provided by experts in the domain, such as epidemiologists working on epidemic modelling. 
\end{itemize}

The following functions are defined:
\begin{itemize}
    \item $\it{NaiveInfectionRate}_{t, \nu}$ takes a history sequence $h \in \Hist$ and computes the infection rate at time $t$ over `the set of nodes $\nu \subseteq V$ as the observed infection rate plus a constant multiplied by the number of nodes observed as unknown.
    \item $\it{InfectionRateOfChange}_{t, \nu}$ takes a history sequence $h \in \Hist$ and computes the change in infection rate between timesteps $t-1$ and $t$ over the set of nodes $\nu \subseteq V$. 
    \item $\it{PercentAction}_{a, N}$ takes a history and returns the percentage of time action $a$ was selected in the last $N$ timesteps.
    \item $\it{ActionSequenceIndicator}_{a_{1:k}}$ is an indicator function returning 1 if the last $k$ actions performed match $a_{1:k}$ and 0 otherwise.
    \item $\it{MAReward}_w$ takes a history and returns the moving average of the reward over a window of size $w$. 
    \item $\it{RateOfChange}$ takes two real numbers and computes the ratio between them.
    \item $\it{ParticleFilter}_{\theta, M}$ takes a history sequence and approximates the belief state using the transition and observation models given by $\theta$ and $M$ particles.
    \item $\it{ParticleInfRate}$ takes a belief state represented by a set of particles and computes the expected infection rate. 
\end{itemize}

The agent has access to the following predicate functions:
\begin{itemize}
    \item $\it{NaiveInfectionRate}_{t, \nu} \comp \it{Encode}_5  \comp \it{Bit}_i \comp (=1) (h) \;\;\; \text{for various } \;i, \nu \subseteq V, \; t = 1$
    \item $\it{InfectionRateOfChange}_{t, \nu} \comp  \it{Encode}_{7}  \comp \it{Bit}_i \comp (=1) (h) \;\;\; \text{for various} \;i, \nu \subseteq V, \; t = 1$
    \item $\it{PercentAction}_{a, N} \comp \it{Encode}_{8}  \comp \it{Bit}_i \comp (=1) (h) \;\;\; \text{for all } a \in \A  \text{ and various values of } N$
    \item $\it{ActionSequenceIndicator}_{a_{1:k}}(h)$ for various $a_{1:k}$, 
 ${k \geq 1}$
    \item $\lambda s.\it{RateOfChange}(\it{MAReward}_{w_1}(s), $ $\it{MAReward}_{w_2}(s)) \comp $ ${( \geq 1) (h)}$ \text{for } $w_1, w_2 \in \N$
    \item $\it{ParticleFilter}_{\theta, M} \comp \it{ParticleInfRate} \comp \it{Encode}_5  \comp \it{Bit}_i \comp$ ${(=1)(h)}$ for various $\theta, M = 100$
\end{itemize}

The $\Phi$-BCTW models introduced in this domain are constructed from a set of $20$ predicates. 

\textbf{Uninformative Predicates.}
The set of uninformative predicates contains predicates that provide limited utility in representing the environment. The most basic uninformative function we generate is $\it{RandomBit}_{p}$ which returns $1$ with probability $p$ and $0$ otherwise. We also consider a predicate function $\it{Randomize}_{p}$ which takes a binary symbol as input and outputs the same symbol with probability $p$. $\it{Randomize}_{p}$ is used to produce noised versions of the informative predicates; given a predicate function $f: \Hist \to \{0, 1\}$, $\it{Randomize}_{p}$ can be used to define a predicate function $f \comp \it{Randomize}_{p}(h)$. Whilst randomized versions of the informative predicates are not strictly `uninformative' (depending on the level of randomization), the level of utility they provide is typically lower than their non-randomized counterparts.

\subsection{Agent Design}
Table \ref{tab:BCTW_config} details the agent configuration used in each domain. The $\abs{M_t}$ column denotes the maximum number of specialists that are available at any one time. 
The $d$ column denotes the number of predicates used to construct the $\Phi$-BCTW model and consequently, the depth of the $\Phi$-BCTW tree. 
The $\epsilon$ column denotes the starting value of $\epsilon$ used in $\epsilon$-greedy exploration and the decay factor column denotes the rate at which the value of $\epsilon$ decreases after each time step. 
Finally, the $H$ column denotes the horizon used to compute the value function and MCTS simulations denotes the number of MCTS simulations computed per time step.

\begin{center}
    \begin{table}[H]
        \centering
        \begin{tabular}{cccc}
         & Biased RPS & Taxi & Epidemic Control\\
        \hline
        $\abs{M_t}$ & 10 & 10 & 10\\
        $d$ & 2 & 17 & 20 \\
        $\epsilon$ & 0.999 & 0.999 & 0.9999\\
        Decay factor & 0.9999 & 0.9999 & 0.999999\\
        $H$ & 4 & 14 & 10\\
        MCTS Sims & 40 & 50 & 20
        \end{tabular}
        \caption{DynamicHedgeAIXI agent learning configuration for each domain.}
        \label{tab:BCTW_config}
    \end{table}
\end{center}

\section{SEIRS epidemic process on contact networks}
\label{appendix:SEIRS_dynamics}

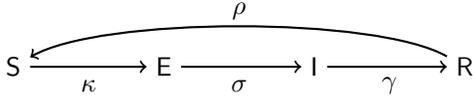
\begin{figure}[H]
    \centering
    \begin{tikzpicture} 
        \node (s2) at (6,0) {$\sS$};
        \node (e2) at (8,0) {$\sE$};
        \node (i2) at (10,0) {$\sI$};
        \node (r2) at (12,0) {$\sR$};
        \draw [->,thick] (s2) to (e2) node[draw=none] at (7,-.25) {$\kappa$};
        \draw [->,thick] (e2) to (i2) node[draw=none] at (9,-.25) {$\sigma$};
        \draw [->,thick] (i2) to (r2) node[draw=none] at (11,-.25) {$\gamma$};
        \draw [->,thick,bend right] (r2) to [looseness=.5] (s2) node[draw=none] at (9, .75) {$\rho$};
    \end{tikzpicture}
    \caption{SEIRS compartmental model on contact networks with transmission rate $\kappa = \frac{1-(1-\beta)^{k_t}}{\omega}$ (where $\beta$ is the contact rate), latency rate $\sigma$, recovery rate $\gamma$ and loss of immunity rate $\rho$. }
    \label{fig:epi}
\end{figure}

We model the epidemic control problem as a stochastic Susceptible-Exposed-Infected-Recovered-Susceptible (SEIRS) process evolving over a contact network.
A contact network is an undirected graph where the nodes represent individuals and the edges represent interactions between individuals. Each individual node in the graph is labelled by one of four states corresponding to their infection status: Susceptible ($\sS$), Exposed ($\sE$), Infectious ($\sI$), and Recovered ($\sR$). Every node also maintains an immunity level $\omega$. The model evolves as a POMDP. At time $t$, the temporal graph is given by $G_t = (V, \mathcal{E}_t)$ with $V$ as the set of nodes and $\mathcal{E}_t$ as the set of edges at time $t$. A function $\zeta_t : V \to \{\sS, \sE, \sI, \sR\} \times \{1, \eta_1, \eta_2\}$ maps each node to its label and one of three immunity levels where $\eta_1, \eta_2 \in  \R_+$. Together, $(G_t, \zeta_t)$ constitute the state. 
At time $t$, a Susceptible node with $k_t$ Infectious neighbours becomes Exposed with probability $\frac{1-(1-\beta)^{k_t}}{\omega}$, where $\beta \in [0, 1]$ is the rate of transmission. An Exposed node becomes Infectious at rate $\sigma$. Similarly, an Infectious node becomes Recovered at a rate $\gamma$ and becomes Susceptible again at a rate $\rho$.

The agent performs an action $a_t$. Quarantine actions modify the underlying connectivity of the graph $G_t$ by removing any edges that contain a quarantined node; this leads to the graph at time $t+1$, $G_{t+1}$. Vaccinate actions modify $\zeta_{2, t}$ such that vaccinated nodes have updated immunity levels in $\zeta_{2, t+1}$. The infection status label of every node evolves according to the following equations:
\begin{equation}
\begin{split}
&\tau(\zeta_{1, {t+1}}(v) | \zeta_{1, t}(v), s_t ) \\
&= 
\begin{cases}
    \frac{1 - (1 - \beta)^{k_t}}{\zeta_{2, t}(v)} \;\; & \text{if } \zeta_{1, {t+1}}(v) = E \text{ and }\zeta_{1, t}(v) = S\\
    1 - \frac{1 - (1 - \beta)^{k_t}}{\zeta_{2, t}(v)} \;\; & \text{if } \zeta_{1, {t+1}}(v) = S \text{ and }\zeta_{1, t}(v) = S\\
    \sigma \;\; & \text{if } \zeta_{1, {t+1}}(v) = I \text{ and }\zeta_{1, t}(v) = E\\
    1 - \sigma \;\; & \text{if } \zeta_{1, {t+1}}(v) = E \text{ and }\zeta_{1, t}(v) = E\\
    \gamma \;\; & \text{if } \zeta_{1, {t+1}}(v) = R \text{ and }\zeta_{1, t}(v) = I\\
    1 - \gamma \;\; & \text{if } \zeta_{1, {t+1}}(v) = I \text{ and }\zeta_{1, t}(v) = I\\ 
    \rho \;\; & \text{if } \zeta_{1, {t+1}}(v) = S \text{ and }\zeta_{1, t}(v) = R\\
    1 - \rho \;\; & \text{otherwise,}
\end{cases}
\label{eqn:SEIRS_transition}
\end{split}
\end{equation}
where $k_t$ denotes the number of Infectious, connected neighbours that $v$ has at time $t$.

The observations resemble the testing for an infectious disease with positive $+$ and negative $-$ outcomes. Recall that the observations on each node are from $\Obs = \{+, -, ?\}$ where $?$ indicates the corresponding individual has unknown/untested status. At time $t$, node $v \in V$ emits an observation according to the following distribution:
\begin{equation}
\begin{split}
    & \xi_t^{v}(+ | \zeta_{1, t}(v) = S) = \alpha_S \mu_S, \\
    & \xi_t^{v}(+ | \zeta_{1, t}(v) = E) = \alpha_E \mu_E, \\
    & \xi_t^{v}(- | \zeta_{1, t}(v) = S) = \alpha_S (1 - \mu_S), \\
    &\xi_t^{v}(- | \zeta_{1, t}(v) = E) = \alpha_\sE (1-\mu_E), \\
    & \xi_t^{v}(? | \zeta_{1, t}(v) = S) = 1 - \alpha_S, \\
    &\xi_t^{v}(? | \zeta_{1, t}(v) = E) = 1 - \alpha_E, \\
    & \xi_t^{v}(+ | \zeta_{1, t}(v) = I) = \alpha_I \mu_I, \\
    &\xi_t^{v}(+ | \zeta_{1, t}(v) = R) = \alpha_R \mu_R, \\
    & \xi_t^{v}(- | \zeta_{1, t}(v) = I) = \alpha_I (1-\mu_I), \\ 
    &\xi_t^{v}(- | \zeta_{1, t}(v) = R) = \alpha_R (1 - \mu_R), \\
    & \xi_t^{v}(? | \zeta_{1, t}(v) = I) = 1 - \alpha_I, \\
    & \xi_t^{v}(? | \zeta_{1, t}(v) = R) = 1 - \alpha_R.
\end{split}
\label{eqn:SEIRS_obs}
\end{equation}

Here, $\alpha_S, \alpha_E, \alpha_I, \alpha_R$ denote the fraction of Susceptible, Exposed, Infectious and Recovered individuals, respectively, that are tested on average at each time step. The parameters $\mu_S, \mu_E, \mu_I, \mu_R$ denote the probability that a node that is Susceptible, Exposed, Infectious, Recovered respectively tests positive.
Table \ref{tab:trans_obs_params} lists the transition and observation model parameters that were used.

\begin{center}
    \begin{table}[H]
        \begin{center}
        \begin{tabular}{cc}
        Parameter & Value\\
        \hline
        $\beta$ & 0.2\\
        $\sigma$ & 0.3\\
        $\gamma$ & 0.08\\
        $\rho$ & 0.1\\
        $\alpha_S$ & 0.1\\
        $\alpha_E$ & 0.1\\
        $\alpha_I$ & 0.8\\
        $\alpha_R$ & 0.05\\
        $\mu_S$ & 0.1\\
        $\mu_E$ & 0.9\\
        $\mu_I$ & 0.9\\
        $\mu_R$ & 0.1\\
        \end{tabular}
        \caption{Transition and observation model parameters.}
        \label{tab:trans_obs_params}
        \end{center}
    \end{table}
\end{center}

\fi

\end{document}